\documentclass[conference]{IEEEtran}
\usepackage{graphicx}
\usepackage{latexsym}
\makeatletter
\@ifpackageloaded{hyperref}{
  \PackageWarning{mypackage}{hyperref is already loaded}
}{
  \usepackage[pagebackref,breaklinks,colorlinks]{hyperref}
}
\@ifpackageloaded{prettyref}{
  \PackageWarning{mypackage}{prettyref is already loaded}
}{
  \usepackage{prettyref}
}
\@ifpackageloaded{cite}{
  \PackageWarning{mypackage}{cite is already loaded}
}{
  \usepackage{cite}
}
\usepackage{booktabs}
\usepackage{cellspace}
\makeatother
\usepackage{amsmath}
\usepackage{amsfonts}
\usepackage{amssymb}
\usepackage{amsthm}
\usepackage{multirow}
\usepackage{algorithmic}
\usepackage{algorithm}
\usepackage{array}
\usepackage{textcomp}
\usepackage{stfloats}
\usepackage{verbatim}
\usepackage{graphicx}
\usepackage{mathtools}

\usepackage{longtable}

\usepackage{xcolor}
\definecolor{correctacc}{RGB}{105,169,203}
\definecolor{misclassified}{RGB}{213,139,71}
\definecolor{mean}{RGB}{96,153,102}

\newcommand{\pref}[1]{\prettyref{#1}}

\newrefformat{sec}{Sec.~\ref{#1}}
\newrefformat{apx}{Appendix~\ref{#1}}
\newrefformat{lem}{Lemma~\ref{#1}}
\newrefformat{fig}{Fig.~\ref{#1}}
\newrefformat{alg}{Algorithm~\ref{#1}}
\newrefformat{table}{Table~\ref{#1}}

\def\p{\boldsymbol{p}}

\def\shape{\boldsymbol{\beta}}
\def\pose{\boldsymbol{\theta}}

\usepackage{cuted}
\usepackage{capt-of}
\def\BibTeX{{\rm B\kern-.05em{\sc i\kern-.025em b}\kern-.08em
    T\kern-.1667em\lower.7ex\hbox{E}\kern-.125emX}}
\hypersetup{
     colorlinks   = true,
     citecolor    = blue
}

\usepackage{fancyhdr}
\usepackage{balance}       

\begin{document}

\title{Generative Learning as a Tool to Improve Perception of Emotional Body Motion Expressions} 

\author{
\IEEEauthorblockN{Huakun Liu}
\IEEEauthorblockA{\textit{Nara Institute of Science and Technology}\\
liu.huakun.li0@is.naist.jp}
\\
\IEEEauthorblockN{Felix Dollack}
\IEEEauthorblockA{\textit{Nara Institute of Science and Technology}\\
felix.d@is.naist.jp}
\\
\IEEEauthorblockN{Chia-huei Tseng}
\IEEEauthorblockA{\textit{Tohoku University}\\
tseng@riec.tohoku.ac.jp}
\and
\IEEEauthorblockN{Miao Cheng}
\IEEEauthorblockA{\textit{Tohoku University}\\
cheng.miao.c3@tohoku.ac.jp}
\\
\IEEEauthorblockN{Victor Schneider}
\IEEEauthorblockA{\textit{Tohoku University}\\
schneider.victor.pierre.d3@tohoku.ac.jp}
\\
\IEEEauthorblockN{Yoshifumi Kitamura}
\IEEEauthorblockA{\textit{Tohoku University}\\
kitamura@riec.tohoku.ac.jp}
\and
\IEEEauthorblockN{Xin Wei}
\IEEEauthorblockA{\textit{Nara Institute of Science and Technology}\\
wei.xin.wy0@is.naist.jp}
\\
\IEEEauthorblockN{Hideaki Uchiyama}
\IEEEauthorblockA{\textit{Nara Institute of Science and Technology}\\
hideaki.uchiyama@is.naist.jp}
\\
\IEEEauthorblockN{Monica Perusquia-Hernandez}
\IEEEauthorblockA{\textit{Nara Institute of Science and Technology}\\
perusquia@ieee.org}
}

\maketitle

\thispagestyle{fancy}

\begin{abstract}
Emotional body motion expressions are an essential element of non-verbal communication.
Effectively conveying these expressions through technology is of utmost importance, for example, with virtual reality avatars and in social robotics.
Recent advances in generative models have opened new opportunities for advancing research on emotional body motion learning.
However, generating accurate emotional expression representations is challenging, given the subtlety of emotional cues, individual variability, and cultural differences.
We investigate whether a generative model can implicitly learn emotional body motions directly from culturally grounded motion-capture data, without explicit emotion-motion guidance.
Using a dataset of emotional performances by 49 Japanese actors, we trained a Transformer-based generative model to generate expressive motions conditioned on 13 discrete emotion labels. 
We evaluate the generated motions from two perspectives: (1) an LSTM-based classifier to assess recognizability by machine observers, achieving a recognition accuracy of 22.80\%, and (2) a human perception study with Japanese raters to assess alignment with human affective interpretations, yielding a recognition accuracy of 24.91\%.
Beyond these, we evaluate the utility of generative modeling for three practical tasks: augmenting emotion recognition models, extracting representative emotion-specific motion patterns, and synthesizing smooth transitions between emotion intensities.
Our findings highlight the potential of implicit, data-driven generative modeling to enhance affective computing applications and our understanding of emotion expressions.
\end{abstract}


\section{Introduction}
\begin{figure}
    \centering
    \includegraphics[width=0.85\linewidth]{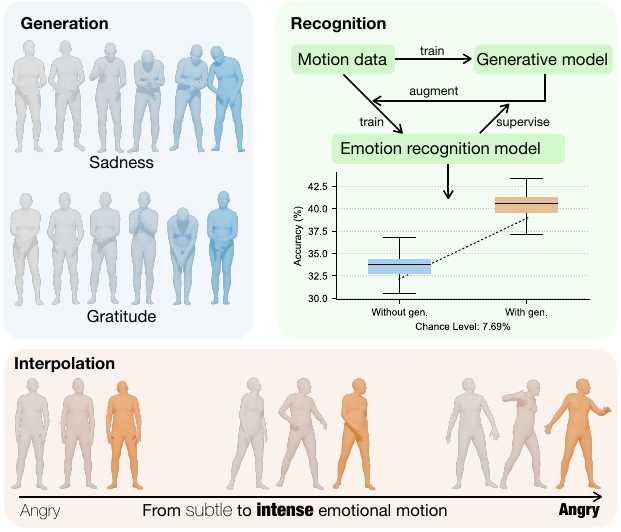}
    \caption{Overview of our study on emotion-conditioned motion generation and its implications for emotion recognition and interpolation. Left: The generative model produces whole-body emotional motions from discrete emotion labels, capturing expressive cues such as the forward-leaning posture in gratitude emotion. Middle: Generated motions, supervised by an emotion recognition model trained on real data, are used to augment the training set, resulting in improved recognition accuracy over using real data alone. Right: The latent space supports smooth interpolation between intensity levels within the same emotion, enabling fine-grained control over expressive variations.}
    \label{fig:teaser}
\end{figure}
Emotion expressions have been investigated primarily for facial and vocal signals, while body motion affective expressions remain under-investigated~\cite{fasel2003automatic, scherer2003vocal, li2020deep}.
Body motion is vital in conveying affect, especially in non-verbal or physically distant communication.
This is increasingly important in applications such as virtual reality (VR), social robotics, and telepresence systems, where full-body motion is a critical channel for interaction.
Generating body motions that express emotions is crucial for creating engaging and naturalistic user experiences and may encourage richer nonverbal behavior than traditional face-focused video platforms~\cite{de2015perception, elansary2024survey}.
Beyond animation and interactive systems, generating emotional body motions also provides a compelling opportunity to study affective motions.
Generative models can uncover core expressive patterns and improve embodied emotion expression recognition by providing data augmentation~\cite{mousavi2025synthetic}.

Motion generation has become increasingly popular, achieving impressive results in tasks such as action synthesis and text-to-motion generation~\cite{petrovich21actor, petrovich2023tmr, jiang2023motiongpt}.
However, emotional body motion generation, which aims to synthesize whole-body motions to convey emotions, remains underexplored.
Emotion expressions tend to be context-dependent~\cite{barrettEmotionalExpressionsReconsidered2019a}, and highly variable among individuals due to personal expressive style~\cite{friedmanPersonalityEnactmentEmotion1980}.
Furthermore, physical conditions, cultural norms, and individual expectations complicate the production and perception of emotional motion~\cite{6212434}.
This makes the mapping from emotion to body motion inherently challenging.

Previous studies addressing emotional body motion generation typically utilize explicit emotional cues and structured supervision to guide the generative process, relying heavily on manually defined emotion-motion relationships.
Prior work mapped particular emotional states to specific limb movements, such as associating sadness with a lowered head posture and a slightly bent torso, and subsequently injecting these handcrafted associations into human limb generations~\cite{yu2024towards}.
While effective in controlled scenarios, such approaches do not capture nuanced expressions of emotion.

We explored the potential of implicitly generating emotionally expressive human motions directly from culturally grounded motion data.
Rather than relying on predefined mappings or handcrafted emotion-to-motion rules, we investigate whether a generative model trained solely on expressive performances can learn meaningful motion patterns associated with emotion categories.
To this aim, we leverage a rich motion dataset of acted emotional performances by 49 Japanese actors, explicitly capturing both individual variability and Japanese-specific cues.
We adopt a Transformer-based variational autoencoder (VAE) to learn a latent representation of emotion-conditioned motions.
During generation, latent vectors are sampled from each emotion category's learned distribution and decoded into pose sequences represented by a human body model.
We evaluate the generated motions through two complementary perspectives: (1) an LSTM-based emotion classifier trained to predict emotion labels to quantify the model's ability to generate emotional patterns; and (2) a human perception study with Japanese raters to examine whether the synthesized motions align with how emotional lay observers interpret expressions.
Furthermore, we explore the practical utility of the generative model in three tasks (\pref{fig:teaser}):\\
\noindent\textbf{1. Data augmentation for emotion recognition.}
Recognizing emotions from human body motion presents challenges due to the inherent variability in expressive styles, limited labeled data, and the subtlety of emotion-specific cues compared to more explicit control signals like action labels.
Generative models offer a promising solution by synthesizing anonymous, emotionally expressive motions that can be used to both evaluate recognition systems and augment training datasets.\\
\noindent\textbf{2. Extraction of representative motion patterns.}
While emotional expressions vary across individuals, we hypothesize that the generative model encodes shared, prototypical features within each emotion category.
By decoding central latent vectors, we aim to uncover common motion tendencies that may support behavioral analysis or synthesis.\\
\noindent\textbf{3. Interpolation across emotional intensities.} Emotion expressions often vary in intensity.
Leveraging the continuity of the learned latent space, we test whether smooth transitions can be generated between different-intensity emotional expressions, offering a way to model graded affective behavior.\\
Through these analyses, our study clarifies both the opportunities and limitations of implicit, data-driven generative modeling as a tool for advancing affective science research on emotional body motion understanding.

\section{Related Works}
\subsection{Motion generation}
Recent advances in motion generation have led to diverse approaches for synthesizing human motions from structured inputs such as action categories and text descriptions.
Early work used recurrent networks for pose prediction~\cite{fragkiadaki_recurrent_2015}, while others improved temporal coherence with hierarchical RNNs~\cite{martinez_human_2017}.
Later, a temporal VAE was used to generate 3D human motions from action labels~\cite{guo2020action2motion}.
Building on this, an action-conditioned transformer VAE (ACTOR) was proposed to sample from a sequence-level latent space conditioned on action labels and duration~\cite{petrovich2021action} with benefits for motion denoising and action recognition. 
Using another approach, Tevet et al.~\cite{tevet2022human} employ a classifier-free transformer-based diffusion model that operates in joint space.
Later, performing diffusion in the motion latent space was proposed to reduce computational overhead while preserving generation quality, resulting in more efficient conditional generation~\cite{chen2023executing}.
In parallel to action categories inputs, efforts in text-to-motion synthesis map language embeddings to motion~\cite{ahuja_language2pose_2019, petrovich2022temos, zhang_we_2021, guo2022generating, jiang2023motiongpt} with a focus on action semantics.
We adopted ACTOR as our backbone generation model, which has been widely used across various motion generation tasks~\cite{petrovich2022temos, petrovich2023tmr}, and offers a strong balance between temporal coherence, computational efficiency, and controllability in sequence-level synthesis~\cite{petrovich21actor}.

\subsection{Emotional motion generation}
Emotion expression has typically been studied through facial~\cite{li_deep_2022} and vocal~\cite{al-dujaili_speech_2023} modalities, leading to early success in generating emotion-aware outputs across multiple modalities~\cite{ferrari_x2face_2018, bhattacharya2021text2gestures,liu2022beat, goyal_emotionally_2023,qi2024emotiongesture,qi2024weakly}.
In contrast, the generation of whole-body emotional motion remains relatively underexplored, despite evidence that emotions can be conveyed through body movement alone~\cite{niedenthal_embodying_2007, tracy_spontaneous_2008}.
Early rule-based approaches map body features to emotional states~\cite{de_meijer_contribution_1989}, but they limit expressive richness by failing to capture the wide variability in how emotions are physically expressed across individuals, contexts, and cultures.
Recent work has attempted to address the data representation heterogeneity and scarcity by leveraging large language models (LLMs).
Emotion-rich textual prompts were used (e.g., ``a man, filled with sadness, walks forward'') and a fine-tuned LLM to infer how emotional states influence specific body parts~\cite{yu2024towards}.
Importantly, emotion-to-limb mappings are manually defined in advance and used as supervision signals during LLM training. These purely rule-based mappings include terms such as ``head: lowered, looking downward'' or ``torso: slightly bent.''
However, the mix of data approach and rule-based methods may suffer from a lack of flexibility, notably with how cultural background and individual differences may shape emotional perception and execution~\cite{elfenbein_universality_2002,kleinsmith_cross-cultural_2006}.
Therefore, we explored whether a generative model can learn to produce expressive whole-body emotional motion directly from performance data, without relying on manually predefined motion-to-emotion guidance.

\section{Methods}

\begin{figure*}
    \centering
    \includegraphics[width=.8\textwidth]{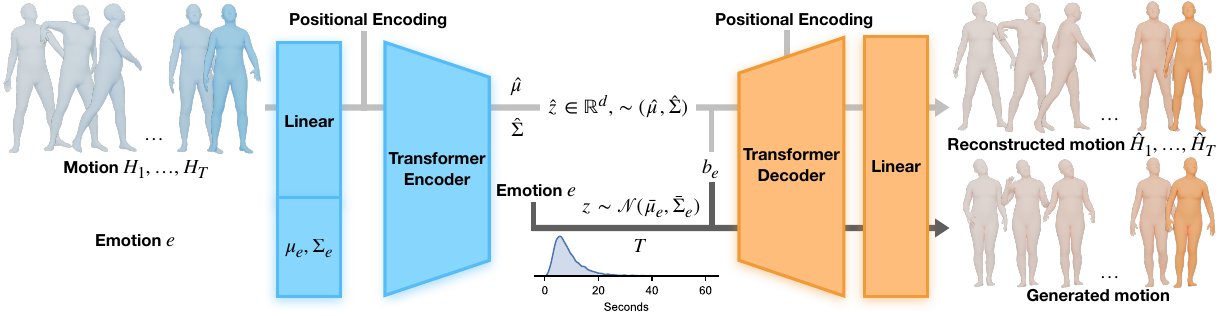}
    \caption{Overview of the ACTOR-based emotional motion generation model~\cite{petrovich2021action}. The model consists of a Transformer-based encoder and decoder trained as a conditional variational autoencoder. The encoder receives a sequence of linearly projected motion inputs $H_1, \ldots, H_T$, concatenated with emotion-specific tokens $\mu_e$ and $\Sigma_e$ corresponding to the given label $e$, and outputs the latent distribution parameters $\hat{\mu}$ and $\hat{\Sigma}$. A latent vector $z$ is sampled and passed to the decoder, along with the emotion embedding $b_e$ and positional encodings, to reconstruct the motion sequence $\hat{H}_1, \ldots, \hat{H}_T$. During inference, generation is performed by randomly sampling $z \sim \mathcal{N}(\bar{\mu}_e, \bar{\Sigma}_e)$ from the learned distribution for emotion $e$.}
    \label{fig:architecture}
\end{figure*}

\subsection{Dataset}
We use a subset of the Diverse Intercultural E-Motion Database of Asian Performers (DIEM-A)~\cite{Cheng2025-uo}, consisting of motion data from 49 Japanese professional performers (27 female, 22 male; mean age = 38.7 years; mean performing experience = 19.6 years).
Each performer was instructed to prepare performances for 13 emotion categories: \textit{joy}, \textit{sadness}, \textit{anger}, \textit{surprise}, \textit{fear}, \textit{disgust}, \textit{contempt}, \textit{gratitude}, \textit{guilt}, \textit{jealousy}, \textit{shame}, \textit{pride}, and \textit{neutral}.
For the 12 non-neutral emotions, performers created three emotion-eliciting scenarios per category, performed at three different intensity levels (low, medium, and high), while the \textit{neutral} emotion required three scenarios without specified intensities. This protocol resulted in 111 motion sequences per performer.
No constraints were imposed on how performers expressed their emotions to ensure diversity in expression styles and motion.

The performances, originally captured using motion tracking, were transformed into the SMPL body model~\cite{loper2015smpl} to represent human motions.
SMPL provides a detailed and expressive mesh-based, surface-level representation of the full human body using two sets of parameters: (1) pose parameters $\pose \in \mathbb{R}^{24 \times 3}$, which define the relative rotations of 23 body joints and a global root orientation in axis-angle format, and (2) shape parameters $\shape \in \mathbb{R}^{10}$, which account for person-specific body shape variations.
The model uses a linear blend skinning function to produce a mesh. To produce realistic motion, the deformed mesh is then posed using joint locations.

To prepare the data for analysis and generation, we first cleaned the raw recordings, downsampled from 120~Hz to 20~Hz, and exported them into C3D format.
These cleaned sequences were then converted into SMPL representations using MoSh++~\cite{AMASS}.
After conversion, we applied a T-pose removal classifier trained on the AMASS dataset~\cite{AMASS, BABEL:CVPR:2021} to detect and remove calibration T-poses and surrounding neutral transitions automatically.
All results were manually verified, and only segments corresponding to the main expressive content of each scenario were retained.
As a result, the final dataset used in this study contains 5,439 motion sequences, totaling approximately 13.6 hours of data.
Sequence durations range from 0.9 to 59.9 seconds (mean = 9.0 s, SD = 6.0 s).

\subsection{Emotion-conditioned motion generation}
\noindent\textbf{Problem definition.}
Our goal is to generate emotionally expressive human body motion, conditioned on a target emotion label.
We focus solely on pose generation, assuming a fixed average body shape.
Formally, given a discrete emotion label $e \in \mathcal{E}$ from a predefined set of 13 categories, we generate a sequence of motion parameters $H_{1:T} = (\pose_t, \p_t)_{t=1}^{T}$, where $\pose_t$ and $\p_t$ denote the pose parameters and root joint translations at time step $t$, respectively.
The sequence length $T$ is randomly sampled from an emotion-specific duration distribution.
Since emotional expression varies between individuals and is culturally influenced, we only used data from Japanese performers.

\noindent\textbf{Generation model.}
As shown in~\pref{fig:architecture}, we adopt ACTOR~\cite{petrovich2021action}, a Transformer-based conditional variational autoencoder (VAE), to generate human motion sequences conditioned on discrete emotion labels.
Unlike autoregressive methods that generate poses frame-by-frame~\cite{hou2020soul}, ACTOR samples a latent vector representing the entire motion sequence from a latent distribution and generates the full motion in a single forward pass.
This design improves computational efficiency and reduces error accumulation over time.
The architecture consists of Transformer-based encoder and decoder modules.
The encoder takes sequences of parameters $H_{1:T}$ along with their corresponding emotion labels as input, encoding them into latent distribution parameters $(\hat{\boldsymbol{\mu}}, \hat{\boldsymbol{\Sigma}})$.
Emotion-specific learnable tokens $\boldsymbol{\mu}_e, \boldsymbol{\Sigma}_e$ explicitly condition the encoder, capturing the emotion variability into the latent space.
The decoder reconstructs motion sequences from a latent vector $\hat{\boldsymbol{z}}$ sampled from this latent distribution.
Specifically, the decoder receives the latent vector, the emotion-specific embedding $\boldsymbol{b}_e$, and sinusoidal positional encodings representing the desired sequence duration as inputs.
The output is a reconstructed sequence $\hat{H}_{1:T}$, which can be directly converted into SMPL body meshes for visualization and further evaluation.

During generation, given an emotion label $e$, a target sequence duration $T$ is sampled from the emotion-specific duration distribution.
The model then selects the corresponding emotion embedding $\boldsymbol{b}_e$ to condition the latent vector $\boldsymbol{z}\sim\mathcal{N}(\bar{\boldsymbol{\mu}}_e, \bar{\boldsymbol{\Sigma}}_e)$, where $(\bar{\boldsymbol{\mu}}_e, \bar{\boldsymbol{\Sigma}}_e)$ represent the average latent distribution parameters computed from all training data labeled with emotion $e$.
This latent vector, combined with the emotion embedding $\boldsymbol{b}_e$ and positional encodings, is fed to the decoder to generate the final motion sequence $\hat{H}_{1:T}$.

\section{Evaluation and Discussion}
\begin{table*}[t]
    \small
    \centering
    \caption{Quantitative evaluation of model variants with and without vertex loss $\mathcal{L}_v$.}
    \begin{tabular}{cccccccc}
        \toprule
        & \multicolumn{2}{c}{Reconstruction} & \multicolumn{3}{c}{Generation} & \multicolumn{2}{c}{Recognition} \\
        \cmidrule(lr){2-3}\cmidrule(lr){4-6}\cmidrule(lr){7-8}
         & Angular Err.\textsuperscript{deg} & Mesh/Joint Err.\textsuperscript{cm} & FID$\downarrow$ & Diversity$\rightarrow$ & Multimodality$\uparrow$ & Accuracy\textsuperscript{\%}$\uparrow$ & Accuracy (Aug. 2000)\textsuperscript{\%}$\uparrow$ \\
        \midrule
         \multirow{2}{*}{Real} & \multirow{2}{*}{---} & \multirow{2}{*}{---} & \multirow{2}{*}{0.07\textsuperscript{$\pm$0.20}} & \multirow{2}{*}{6.30\textsuperscript{$\pm$0.13}} & \multirow{2}{*}{3.08\textsuperscript{$\pm$0.02}} & \multirow{2}{*}{34.28\textsuperscript{$\pm$0.18}} & 42.87\textsuperscript{$\pm$1.08} (w/o $\mathcal{L}_v$) \\
         & & & & & & & 41.21\textsuperscript{$\pm$0.59} (w $\mathcal{L}_v$) \\
        w/o $\mathcal{L}_v$ & \textbf{6.31\textsuperscript{$\pm$4.78}} & \textbf{3.14\textsuperscript{$\pm$2.90}}/\textbf{2.52\textsuperscript{$\pm$2.55}}& 0.28\textsuperscript{$\pm$0.02} & 6.00\textsuperscript{$\pm$0.09} & \textbf{4.99\textsuperscript{$\pm$0.03}} & \textbf{23.35\textsuperscript{$\pm$1.65}} & \textbf{30.28\textsuperscript{$\pm$1.45}} \\
        w $\mathcal{L}_v$ & 6.49\textsuperscript{$\pm$4.91} & 3.24\textsuperscript{$\pm$3.00}/2.60\textsuperscript{$\pm$2.62} & \textbf{0.25\textsuperscript{$\pm$0.02}} & \textbf{6.06\textsuperscript{$\pm$0.08}} & 4.96\textsuperscript{$\pm$0.03} & 22.80\textsuperscript{$\pm$1.85}  & 28.79\textsuperscript{$\pm$1.56} \\
        \bottomrule
    \end{tabular}
    \vspace{1em}
    \label{table:eval}
\end{table*}
We evaluated the generative model from technical and perceptual perspectives to assess its capability in capturing and synthesizing emotionally expressive body motions.
Our analysis covers reconstruction accuracy and generation diversity~(\pref{sec:recon_gen}), emotion recognizability by machine classifiers and human observers~(\pref{sec:recog}), and utility in tasks such as emotion recognition~(\pref{sec:recog}), representative motion extraction~(\pref{sec:common_motion}), and intensity interpolation~(\pref{sec:interp}). 

\subsection{Reconstruction and generation}\label{sec:recon_gen}
We compared two variants of the model: one with the mesh vertex loss term $\mathcal{L}_v$ and one without (see supplementary materials for details), as this is the primary factor influencing reconstruction fidelity and surface-level consistency~\cite{petrovich21actor}.
We evaluated their performance in both reconstruction and generation settings.
For \textit{reconstruction}, we assessed how accurately the model can reproduce input motion sequences, thereby capturing the details of observed motions.
For \textit{generation}, we analyzed the quality and diversity of motions sampled from the learned latent space, conditioned only on emotion labels.
The focus of our evaluation is whether the model can generate diverse, semantically consistent motions and how such data might be used in emotion-related tasks.

For the \textbf{reconstruction evaluation}, we computed three metrics: \textit{angular error}, which measures the difference in joint rotations between the reconstructed and ground-truth poses; \textit{mesh error}, and \textit{joint error}, calculated as the Euclidean distance between corresponding vertices and joints of the reconstructed and ground-truth SMPL body meshes.
For \textbf{generation evaluation}, we checked the realism of sampled motions using the \textit{Fréchet Inception Distance (FID)}, which compares the distribution of generated motions to that of real motions in a learned feature space.
To compute FID, we adopted an RNN-based emotion recognition model trained on the real motion dataset. We extracted feature embeddings from the model's final layer and computed FID between the feature distributions of 2,600 randomly sampled real and generated motions. This process was repeated 20 times and averaged to ensure robust estimation. To promote statistical stability, we computed FID jointly across all emotion categories.
A lower FID indicates greater similarity to real data. 
We further assessed the variability of generated motions using two metrics: diversity and multimodality. For reference, we also computed these metrics on real motion data.
\textit{Diversity} is measured as the average pairwise feature distance between randomly sampled motion sequences, reflecting the overall variation across the generation space.
Values closer to those of real data are preferred.
\textit{Multimodality} quantifies the average feature distance among multiple samples conditioned on the same emotion label, capturing the model's ability to express within-class variability. Higher values indicate better within-category variation.
To capture semantically meaningful variations, we computed generation metrics only on generated samples that were correctly classified by the emotion recognition model. This is the same filtered subset used in the one-time supervised augmentation setup~(\pref{sec:recog}) used to mitigate the risk of misleading metric values due to off-target generations and to ensure that measured values reflect expressive variation within an emotion class.
As shown in~\pref{table:eval}, both model variants---one trained with the vertex loss term $\mathcal{L}_v$ and one without---achieve reconstruction quality comparable to those used in action- or text-to-motion generation tasks~\cite{chen2023executing, zhou2023unified}, with mesh errors within 4~cm.
The results reveal no substantial difference between the two variants in emotional body motion generation, suggesting that the vertex loss $\mathcal{L}_v$, while providing stronger surface-level supervision, does not significantly enhance reconstruction accuracy in the emotional model generation task.
Notably, both variants achieved higher multimodality than those computed on real data, indicating that the generative model has a strong capacity to generate diverse expressions within each emotion category.
This highlights its potential for augmenting motion-based emotion recognition tasks.

\subsection{Emotion expression recognition from human body motion}\label{sec:recog}
We investigated whether a generative model can support emotion expression recognition in two complementary roles: (1) generating emotionally expressive motions that are recognizable by machines and humans, and (2) providing additional training data to improve the performance of data-driven recognition models.
To ensure model-agnostic evaluation, we used a unidirectional multi-layer long short-term memory (LSTM)~\cite{hochreiter1997long} recurrent neural network as our recognition model.
This allows us to assess motion discriminability without bias toward any specific architecture.
To train the baseline recognition model, the DIEM-A data was split into 80\% training set, 10\% validation set, and 10\% test set.
The generation model was trained on the same training set, ensuring no data overlap or information leakage during the evaluation.

\noindent\textbf{Machine-based emotion expression recognition on real and generated motions.}
As shown in~\pref{table:eval}, real body motion sequences achieve the highest recognition accuracy at 34.28\%, compared to a chance level of 7.69\%, establishing a baseline for the model's ability to extract affective cues from real body motions.
This also highlights the difficulty of emotion expression recognition from body motion alone in DIEM-A dataset, given the freely performed and highly variable expressive body motions.
In comparison, generated motions achieve an accuracy of 22.80\% on 2,000 randomly generated motions, indicating that the synthesized sequences retain emotion-specific patterns recognizable by a downstream classifier.
However, a noticeable gap of approximately 12\% remains between real and generated data, consistent with prior findings in motion synthesis and recognition~\cite{petrovich21actor}.
This gap may arise from distributional discrepancies, where generated motions partially deviate from the true data distribution and lack cues critical for emotion expression recognition.
\begin{figure*}
    \centering
    \includegraphics[width=\textwidth]{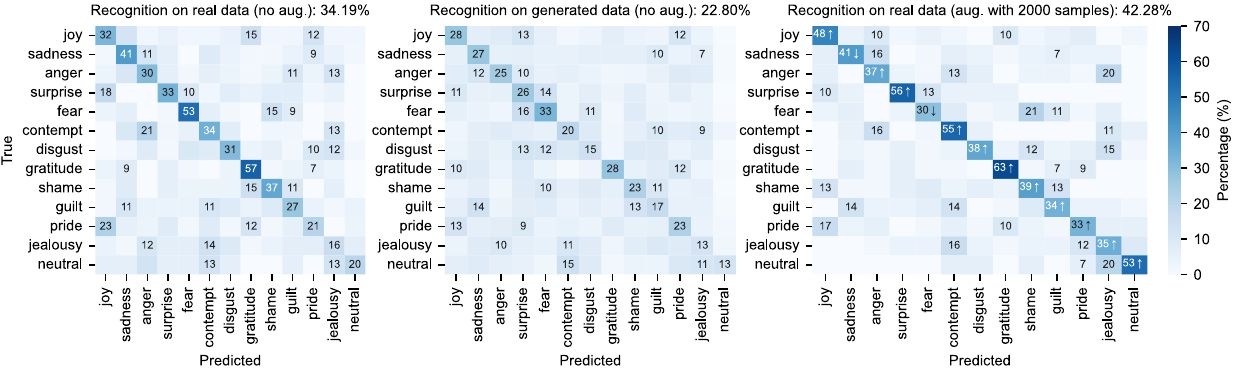}
    \caption{Confusion matrices of the emotion recognition model evaluated without augmentation on real motion data (left), generated motion data (middle), and with 2000 generated samples used for augmentation (right), with top-3 predictions annotated in each row.}
    \label{fig:confusion_matrix}
\end{figure*}
\pref{fig:confusion_matrix} shows the confusion matrix of recognition results.
For real motions, emotion categories such as \textit{fear}, \textit{gratitude}, and \textit{sadness} show relatively high recognition accuracies, because of their distinct and culturally consistent body cues, such as withdrawn or protective motions for fear.
In contrast, emotions with more subtle expressions, such as \textit{shame} and \textit{guilt}, exhibit greater confusion in both real and generated data.
While the confusion patterns are broadly aligned between real and generated motions, predictions on generated data are less concentrated and show reduced classification accuracy.
This highlights the challenge of modeling subtle expressive differences and suggests that further refinement of the generative model is needed to improve the distinctiveness of emotion- and context-specific motion patterns.

\noindent\textbf{One-time vs. iterative augmentation strategies.}
\begin{figure}
    \centering
    \includegraphics[width=0.85\linewidth]{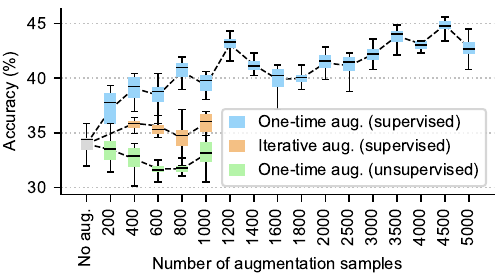}
    \caption{Comparison of recognition accuracy under different data augmentation strategies and sample sizes.}
    \label{fig:augmentation_accuracy}
\end{figure}
We compared two augmentation strategies: one-time and iterative augmentation.
In the \textit{one-time approach}, a fixed number of generated samples per emotion category is selected using the baseline classifier, and added to the training set in a single step.
In the \textit{iterative approach}, the classifier is incrementally retrained from the baseline model, where each new model builds on the previous one by adding newly selected high-confidence samples to the training set.
For example, the model trained with 200 augmented samples is used to select additional data, which is then used to train the next model with 400 samples, and so on.
As shown in~\pref{fig:augmentation_accuracy}, both strategies yield improvements over the no-augmentation baseline.
However, the one-time augmentation consistently outperforms the iterative approach, especially when a sufficient number of samples is available.
This suggests that while iterative selection may introduce additional diversity, it also risks reinforcing early misclassifications or redundant patterns.
By contrast, selecting a batch of synthetic samples using a strong initial classifier leads to more stable and substantial improvements in recognition performance.

\noindent\textbf{Importance of Sample Quality in Augmentation.}
We compared supervised and unsupervised strategies for selecting generated samples to augment the training of emotion expression recognition models. 
The supervised approach filters samples using the baseline classifier, retaining only those correctly classified and thus semantically aligned with the intended emotion.
In contrast, the unsupervised strategy selects samples randomly. 
As shown in~\pref{fig:augmentation_accuracy}, supervised augmentation consistently outperforms the unsupervised one across all augmentation sizes.
The unsupervised method resulted in a limited improvement, suggesting that inaccurate or ambiguous samples introduce noise and degrade recognition model performance.
These results demonstrate the importance of quality control in sample selection and highlight the benefit of leveraging classifier feedback to guide data augmentation.

\noindent\textbf{Effect of the Number of Augmented Samples.}
We evaluated how the quantity of synthetic samples affects recognition performance.
We incrementally added varying numbers of generated samples to the training set and evaluated the resulting classification accuracy.
As shown in~\pref{fig:augmentation_accuracy}, accuracy steadily improves with the number of augmented samples, peaking at 42.3\% with 2,000 generated motions, significantly higher than the no-augmentation baseline of 34.2\%.
Notably, performance begins to plateau beyond 1,200 samples and slightly fluctuates, suggesting a saturation point where additional synthetic data offers diminishing returns or introduces redundant information.
The confusion matrices in~\pref{fig:confusion_matrix} highlight the effect of augmentation on individual emotion categories.
With 2,000 augmented samples, benefiting from the increased diversity and quantity of training samples, improvements are observed in most categories, particularly \textit{joy}, \textit{anger}, \textit{gratitude}, \textit{shame}, and \textit{neutral}, which show clear increases in class-wise accuracy.
These results demonstrate that augmenting with a moderate number of high-quality generated emotional motions enhances recognition performance, particularly in categories with distinctive expressive cues.
However, the improvement becomes increasingly limited as more samples are added, suggesting diminishing returns beyond a certain quantity.

\noindent\textbf{Human-based emotional expression recognition on generated motions.}
\begin{figure}
    \centering
    \includegraphics[width=0.81\linewidth]{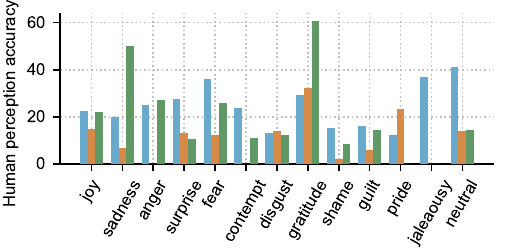}
    \caption{Human perception accuracy of generated motions. The average accuracies for motions \textcolor{correctacc}{correctly classified}, \textcolor{misclassified}{misclassified}, and \textcolor{mean}{decoded from mean latent vectors} are 24.91\%, 10.99\%, and 21.98\%, respectively.}
    \label{fig:human_eval}
\end{figure}
To evaluate the perceptual validity of the generated motions, we conducted a human evaluation study with 20 Japanese raters (16 males, 4 females, mean age 43.5 years, average age closely matches that of the motion performers).
Participants were asked to identify emotions from 65 generated motion sequences (total duration: 7.02 minutes, mean duration per sequence: 6.47 seconds, SD = 3.04 seconds).
Of these sequences, 13 were generated from the mean latent vector for each emotion category~(see~\pref{sec:common_motion}); 26 were correctly classified by the trained recognition classifier; and 26 were misclassified by the classifier.
This setup enabled us to analyze how machine classification confidence correlates with human perception and whether it can serve as a proxy for perceptual quality. 
Participants viewed each motion and selected the most appropriate emotion label from the 13 options, without additional cues.
Completion time was an average of 23.80 minutes.
Correlation analysis revealed no significant relationship between response time and accuracy (Pearson's $r = -0.23$, $p = 0.33$).
\pref{fig:human_eval} summarizes perceptual evaluation results.
Motions correctly classified by the machine classifier achieved the highest human recognition accuracy at 24.91\%.
Conversely, misclassified motions yielded a lower accuracy of 10.99\%, confirming that trained classifier predictions can serve as a reliable indicator for selecting higher-quality synthetic data.
However, the overall accuracy across three conditions was only 18.75\%, demonstrating that generated motions still struggle to convey clearly recognizable expressive cues.
These findings underscore the promise of generative models for synthesizing perceptually meaningful emotional motion data and their current limitations in capturing expressive nuances necessary for reliable human recognition.
To contextualize these results, a recent study~\cite{cheng2024toward} reported 41.6\% human accuracy on real motion data from the same dataset, highlighting the gap between performed and perceived emotion.
This indicates a difference between emotion production and perception, perhaps due to the performer's acting ability.
Therefore, our observed results fall within a realistic and interpretable range, given the broader challenges of expressing emotions through body motion and accurately perceiving that motion as the intended affective message.

\subsection{Common motion extraction}\label{sec:common_motion}
\begin{figure}
    \centering
    \includegraphics[width=.95\linewidth]{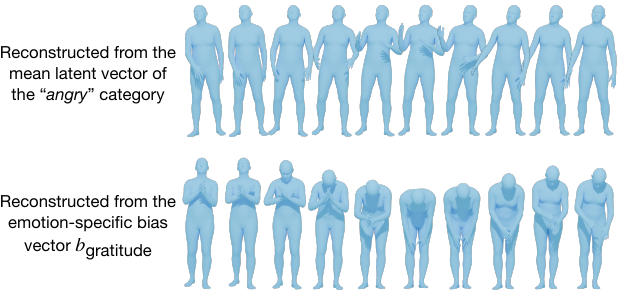}
    \caption{Motion decoded from the mean latent vector computed across all training samples labeled as \textit{angry} (top), and
    from the learned emotion-specific bias vector $b_{\text{gratitude}}$ (bottom).
    }
    \label{fig:common-motion}
\end{figure}
Furthermore, to explore whether the generative model captures shared patterns of emotional expression, we analyzed the latent space by extracting motion sequences that represent common patterns within each emotion category.
We adopted two strategies.
First, we computed the mean latent vector for each emotion category by averaging all latent vectors obtained from training sequences with the same label.
We then decoded each mean vector into a motion sequence, representing the central tendency of the learned distribution.
This approach produces motions that reflect shared characteristics across multiple performers and scenarios.
In particular, individual variations are suppressed while the common motion patterns associated with each emotion are retained.
Second, we decoded from the learned emotion-specific bias vectors $b_e$ used for conditioning the latent distribution.
These vectors act as a distilled representation of each category, independent of any specific input sequence.
By decoding them directly, we obtain a complementary perspective on what the model has implicitly learned as the core features of each emotional expression.

\pref{fig:common-motion} shows two qualitative results for the \textit{angry} and \textit{gratitude} emotions, generated from the mean latent vector and the emotion-specific bias vector, respectively.
We observed that both averaging over expressive motions and learning emotion embeddings tend to suppress individual variations.
However, the decoded motions still retain a few key features that are representative of each emotional category.
For example, the \textit{gratitude} motion displays a forward-leaning posture with hands held close to the chest, which is an expressive cue commonly associated with appreciation in Japanese culture.
These observations suggest that both the mean latent vectors and the learned emotion-specific bias vectors encode semantically meaningful motion priors for each emotion.
\pref{fig:human_eval} shows the results of the human perceptual evaluation on motions decoded from mean latent vectors.
These representative motions achieved a recognition accuracy of 21.98\%, slightly lower than motions correctly classified by the recognition model (24.91\%), but substantially higher than misclassified samples (10.99\%).
Notably, motions for sadness and gratitude yielded relatively high perceptual accuracy, suggesting that these emotions are more consistently expressed across performers.
However, the overall moderate accuracy also suggests that these motions may lack the expressive detail necessary for clearly recognizable expressions of emotion.

\subsection{Emotional motion interpolation}\label{sec:interp}
\begin{figure}
    \centering
    \includegraphics[width=.85\linewidth]{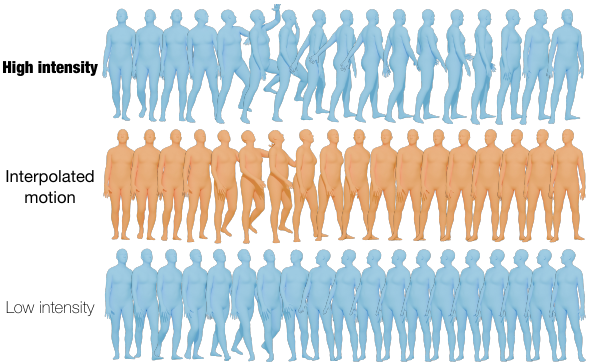}
    \caption{\textit{Angry} motion interpolation between low and high intensity sequences.
    The interpolated sequence exhibits motion characteristics that fall between the two endpoints, demonstrating a smooth transition in intensity.
    }
    \label{fig:interpolation}
\end{figure}
The DIEM-A dataset provides a unique opportunity to explore emotional intensity interpolation, as it includes multiple performances of each emotion expressed at three distinct intensity levels.
Thus, instead of generating new emotional body motion expressions through random sampling, we synthesized new motions by interpolating between latent vectors of motions that share the same emotion label.
We encoded two motion sequences of the same emotion but with different intensities using the trained encoder to obtain their latent representations.
We then computed a new latent vector by taking a weighted average of the two, simulating an intermediate emotional intensity.
The resulting vector is decoded into a motion sequence.
\pref{fig:interpolation} illustrates a qualitative example of emotional motion interpolation between low and high-intensity performances of the ``angry'' emotion.
The top and bottom rows show the source motions with high and low intensity, respectively, while the middle row presents the interpolated motion generated from an average of their latent vectors.
The interpolated motion exhibits a smooth transition in both posture and dynamics.
It clearly reflects characteristics that lie between the subtle, minimal movement of the low-intensity motion and the energetic movement of the high-intensity motion.
This result demonstrates that the model captures a meaningful internal representation of emotional intensity, allowing it to generate expressive motions along a continuous spectrum rather than relying on discrete intensity labels.
Such interpolation ability opens new opportunities for emotion-aware animation control and continuous emotion modeling, enabling more flexible and expressive applications in affective computing and virtual character design.

\section{Conclusions and Future Work}
Our results show the feasibility of generating emotional body motions using a model originally developed for action-based generation. Although the task is challenging, the generated motions improved machine recognition of body expressions of emotion. Machine recognition also supports the pre-selection of body motions that can be used to convey an effective message to human observers, as evidenced by a higher perception recognition of correctly labeled motions by a machine. However, the generated motions still lag behind real data in terms of classification accuracy and perceptual clarity.
Furthermore, generated data are either randomly sampled or filtered using a pre-trained classifier in our current implementation. This static selection may propagate misleading samples.
To address this, future work could incorporate a generative adversarial network to iteratively refine the generated motions by integrating feedback from recognition models or human evaluators in a data-centric loop.

\section*{Ethical impact statement}
This research was a data analysis of the Diverse Intercultural E-Motion Database of
Asian Performers (DIEM-A)~\cite{Cheng2025-uo}. All data provided is anonymous and obtained following the Declaration of Helsinki after obtaining ethical approval.  
Our results have several limitations. First, the emotions are expressed by the actors. Even though the actors were asked to reminisce about a scenario where they would feel a particular emotion, the expressions were posed and not spontaneous~\cite{tian2015emotion}. Therefore, our results need to be interpreted from a perception perspective as a plausible communication message, as opposed to identifying how a person truly feels.
The emotion recognition study involved 20 Japanese raters. The cultural and gender composition of this sample may limit the generalizability of the results. Perception of emotion can be shaped by cultural norms and individual background, and the use of a culturally homogeneous rater group should be considered when interpreting the findings.
Moreover, our models provide baseline results that still need to be improved. Our human rating results show that the human interpretation of an expression might differ from that of a machine recognition model. There is a challenge due to individual differences when expressing an emotion, and this is just the first attempt to find the commonalities in body expressions of emotion.

\section*{Acknowledgment}
This work was supported by the Japan Science and Tech-
nology Agency under the Broadening Opportunities for
Outstanding Young Researchers and Doctoral Students in
Strategic Areas (BOOST) JPMJBS2423 and the RIEC Nation-Wide Cooperative Research Projects Grant Number R05/A33.

\balance{}

\bibliographystyle{IEEEtran}
\bibliography{main.bib}

\end{document}